\documentclass[fleqn]{llncs}
\usepackage{amsmath}

\title{ASHACL:  Alternative Shapes Constraint Language}
\author{Peter F. Patel-Schneider}
\institute{Nuance Communications \\
\email{peter.patel-schneider@nuance.com}}

\begin{document}
\maketitle{}
\thispagestyle{plain}

\begin{abstract}
ASHACL, a variant of the W3C Shapes Constraint Language, is designed to determine
whether an RDF graph meets some conditions.  These conditions are grouped
into shapes, which validate whether particular RDF terms each meet the
constraints of the shape.  Shapes are themselves expressed as RDF triples in an
RDF graph, called a shapes graph.   
\end{abstract}

\section{Introduction}

As of February 2017 the W3C Shapes Constraint Language (SHACL) \cite{shacl}
is under 
development by the W3C Data Shapes Working Group.  
The design document for SHACL, {\em Shapes Constraint Language (SHACL)}, as of
this time is available at 
\url{https://www.w3.org/TR/2017/WD-shacl-20170202/}

\medskip

SHACL is divided into two parts, SHACL Core and SHACL-SPARQL.

SHACL Core has two separate specifications; a complete normative
specification in text and a partial informative specification as a mapping
to an extension of SPARQL \cite{sparql}.  The specification via a mapping to
SPARQL uses a version of
pre-binding defined in SHACL.  The use of pre-binding has caused continual
problems in SHACL.   Sometimes the definition of pre-binding produces many
constructs whose mapping specification diverges from the text specfication.
Sometimes, including now, the definition of pre-binding has problems making
it unsuitable for use in all or part of SHACL.
The text specification of SHACL Core suffers from a lack of rigour,
resulting in a specification that can be read in multiple ways.

SHACL-SPARQL depends on a mapping to SPARQL that has the same problems as the
partial mapping for SHACL Core but the mapping is normative for SHACL-SPARQL.

The syntax of SHACL Core is unnecessarily complex and fragile.  Certain
constructs can be used in some places but not in other very simlar places.
There are constructs that are limited versions of other constructs.  There
are multiple influences on how certain constructs behave.

\medskip

This document contains a specification for ASHACL, an alternative version of
the validating portion of SHACL Core using many parts of the
specification of SHACL Core \cite{shacl}.

The major syntax difference between ASHACL and the current version of SHACL
Core is to unify the various kinds of shapes in SHACL Core into a single notion
of a shape.  The single property {\tt sh:path} then controls how to
determine value nodes.  This results in a simpler and more uniform language.

Another significant difference is how validation results and reports are
defined.  This eliminates the problematic procedural dependencies in SHACL
Core and correctly specifies what needs to be contained in a validation
report.

A third significant difference is the removal of the informative SPARQL
specifications for many SHACL Core constructs.  This eliminates the
informative dependency on pre-binding as well as removing a source of
divergence in the specification.

The specification of ASHACL here is in text form with dependence on only a
few SPARQL functions.  The specification of ASHACL here
has been carefully constructed to be precise and well-defined, eliminating
the problems with imprecise and poorly constructed parts of the current
specification of SHACL.

Because ASHACL is a variant of SHACL, much of the desgin of ASHACL here is
taken fromd SHACL \cite{shacl}.  Some of the wording here has as source
wording in {\em Shapes Constraint Language (SHACL)} \cite{shacl}.

\section{Examples}

Here are several simple informative examples of ASHACL shapes graphs along
with English glosses of what they do.


\noindent
Check that all ASHACL instances of {\tt ex:Person} have all their values for
{\tt ex:child} be ASHACL instances of {\tt ex:Person}.

\begin{verbatim}
@prefix sh:   <http://example.org/ns/shacl\#> .
@prefix ex:   <http://example.org/example/> .
ex:s a sh:Shape ;
  sh:targetClass ex:Person ;
  sh:shape [ sh:path ex:child ;
             sh:class ex:Person ] .
\end{verbatim}

\noindent
Check that all ASHACL instances of {\tt ex:Person} have all their values for
{\tt ex:child} be ASHACL instances of {\tt ex:Person}, using only one shape.

\begin{verbatim}
@prefix sh:   <http://example.org/ns/shacl\#> .
@prefix ex:   <http://example.org/example/> .
ex:s a sh:Shape ;
  sh:targetClass ex:Person ;
  sh:path ex:child ;
  sh:class ex:Person .
\end{verbatim}

\noindent
Check that all subjects and objects of {\tt ex:parent} RDF triples are ASHACL
instances of {\tt ex:Animal}.

\begin{verbatim}
@prefix sh:   <http://example.org/ns/shacl\#> .
@prefix ex:   <http://example.org/example/> .
ex:s a sh:Shape ;
  sh:targetSubjectsOf ex:parent ;
  sh:targetObjectsOf ex:parent ;
  sh:class ex:Animal .
\end{verbatim}

\section{Basic Terminology}

Throughout the remainder of this document, IRIs are written as Turtle
\cite{turtle-1.1} prefixed names, using the following mapping of prefixes to
namespaces:

\begin{tabular}{ll}
{\bf Prefix}	& {\bf Namespace} \\
{\tt xsd}	& {\tt http://www.w3.org/2001/XMLSchema\#} \\
{\tt rdf} 	& {\tt http://www.w3.org/1999/02/22-rdf-syntax-ns\#} \\
{\tt rdfs} 	& {\tt http://www.w3.org/2000/01/rdf-schema\#} \\
{\tt owl}	& {\tt http://www.w3.org/2002/07/owl\#} \\
{\tt sh} 	& {\tt http://example.org/ns/shacl\#}
\end{tabular}

Terminology that is linked to portions of RDF 1.1 Concepts and Abstract
Syntax \cite{rdf11-concepts} is used in ASHACL as defined there.  Terminology that is
linked to portions of SPARQL 1.1 Query Language \cite{sparql} is used in
ASHACL as defined there.  A single linkage is sufficient to provide a
definition for all occurrences of a particular term in this document.

Definitions are complete within this document, i.e., if there is no rule to
make a some situation true in this document then the situation is false.
Definitions, or partial definitions, have the phrase being defined in bold.
Some definitions are dispersed throughout the document.  For these terms
each portion of the definition has the phrase being defined in bold.

\begin{description}

\item[Property:]

A {\bf property} is an
\href{https://www.w3.org/TR/rdf11-concepts/#dfn-iri}{\em IRI}.

\item[Value:]

An
\href{https://www.w3.org/TR/rdf11-concepts/#dfn-rdf-term}{\em RDF term} $n$
has {\bf value} $v$ for property $p$ in an
\href{https://www.w3.org/TR/rdf11-concepts/#section-rdf-graph}{\em RDF graph} $G$ if there is
an
\href{https://www.w3.org/TR/rdf11-concepts/#section-triples}{\em RDF triple} in $G$ with
\href{https://www.w3.org/TR/rdf11-concepts/#dfn-subject}{\em subject} $n$,
\href{https://www.w3.org/TR/rdf11-concepts/#dfn-predicate}{\em predicate} $p$,
and \href{https://www.w3.org/TR/rdf11-concepts/#dfn-object}{\em object} $v$.

An RDF term $n$ has {\bf value} $v$ for
\href{https://www.w3.org/TR/sparql11-query/#pp-language}{\em SPARQL property path expression} 
$p$ in an
RDF graph $G$ if there is a solution mapping in the result of the SPARQL query
{\tt SELECT ?s ?o WHERE \{ ?s $p'$ ?o \}} on $G$ that binds {\tt ?s} to $n$
and {\tt ?o} to $v$, where $p'$ is SPARQL surface syntax for $p$.

\item[Path:]

A {\bf path} in an
RDF graph $G$ 
from
RDF term $n$
to RDF term $m$ is a finite sequence 
of
RDF triples
in $G$ such that the
subject of the first RDF triple is
$n$, 
the
object of the last RDF triple is $m$, and the object of each RDF triple
except the last is the subject of the next.

\item[ASHACL List:]

An {\bf ASHACL list} in an RDF graph $G$ is an IRI or a 
\href{https://www.w3.org/TR/rdf11-concepts/#section-blank-nodes}{\em blank node} that is
either {\tt rdf:nil} (provided that {\tt rdf:nil} has no value for either
{\tt rdf:first} or
{\tt rdf:rest} in $G$), or has exactly one value for {\tt rdf:first} in $G$ and exactly one
value for {\tt rdf:rest} in $G$ that is also an ASHACL list in $G$ and there is
no non-empty path in $G$ from the list back to itself where the predicates of
the RDF triples in the path are each {\tt rdf:rest}.

The {\bf members} of any ASHACL list except {\tt rdf:nil} in an RDF graph $G$ consist
of its value for {\tt rdf:first} in $G$ followed by the members in $G$ of its value
for {\tt rdf:rest} in $G$.  The ASHACL list {\tt rdf:nil} has no {\bf members} in any RDF
graph.

\item[ASHACL Instance and ASHACL Type:]

An RDF term $n$ is an {\bf ASHACL instance} of an RDF term $m$ in an RDF graph $G$ if
there is a path in $G$ from $n$ to $m$ where the predicate of the first RDF triple
in the path is {\tt rdf:type} and the predicates of any other RDF triples in the
path are {\tt rdfs:subClassOf}.

An RDF term $n$ has an RDF term $m$ as {\bf ASHACL type} in an RDF graph $G$ if $n$ is an
ASHACL instance of $m$ in $G$.

\item[Data Graph:]
A {\bf data graph} is any RDF graph.

\end{description}

\section{ASHACL Property Paths}

ASHACL uses an RDF encoding of much of SPARQL property paths.  

A blank node is an {\bf ill-formed property path} in an RDF graph $G$ if it has
a value for more than one of 
either {\tt rdf:first} or {\tt  rdf:rest}, 
{\tt sh:alternativePath}, {\tt sh:inversePath},
{\tt sh:zeroOrMorePath}, {\tt sh:oneOrMorePath}, and {\tt sh:zeroOrOnePath} in $G$.

A blank node is an {\bf ill-formed property path} in an RDF graph $G$ if there is
a path in $G$ from the node back to itself where the sequence of
predicates of the RDF triples in the path matches the regular expression
\begin{align*}
( \; & ( \; \mbox{\tt rdf:rest}* \; \mbox{\tt rdf:first} \; ) \; | \\
     & \mbox{\tt sh:alternativePath} \; ( \mbox{\tt rdf:rest}* \; \mbox{\tt rdf:first} \;) \; | \\
     & \mbox{\tt sh:inversePath} \; | \; \mbox{\tt sh:zeroOrMorePath} \; | \\
     & \mbox{\tt sh:oneOrMorePath} \; | \; \mbox{\tt sh:zeroOrOnePath} \; ) \;\; +
\end{align*}

An RDF term is an {\bf ill-formed property path} in an RDF graph if it
does not satisfy exactly one of the conditions in the mapping below.

The {\bf path mapping} in an RDF graph $G$ of a RDF term $p$ that is an ASHACL
property path in $G$, $path(p,S)$, is a 
\href{https://www.w3.org/TR/sparql11-query/#pp-language}{\em SPARQL property path expression}
defined as follows:
\begin{enumerate}
\item If $p$ is an IRI then $path(p,S)$ is {\sl PredicatePath($p$)}.
\item If $p$ is a blank node that is an ASHACL list in $G$ that has at least two
members in $G$ and none of these members are ill-formed property paths in $G$
then $path(p,S)$ is {\sl SequencePath($path(v_1,S)$ \ldots $path(v_n,S)$)} where $v_i$ are the
members of $p$ in $G$, in order.
\item If $p$ is a blank node that has exactly one value for {\tt sh:alternativePath} in
G and that value is an ASHACL list in $G$ that has at least two members in $G$ and
none of the members are ill-formed property paths in $G$ then $path(p,S)$
is {\sl AlternativePath($path(v1,S)$ \ldots $path(vn,S)$)} where $v_i$ are the members of
the list in $G$, in order.
\item If $p$ is a blank node that has exactly one value $v$ for {\tt sh:inversePath} in $G$
and $v$ is not an ill-formed property path in $G$ then $path(p,S)$ is
{\sl InversePath($path(v,S)$)}.
\item If $p$ is a blank node that has exactly one value $v$ for {\tt sh:zeroOrMorePath}
in $G$ and $v$ is not an ill-formed property path in $G$ then $path(p,S)$ is
{\sl ZeroOrMorePath($path(v,S)$)}.
\item If $p$ is a blank node that has exactly one value $v$ for {\tt sh:oneOrMorePath} in
G and $v$ is not an ill-formed property path in $G$ then $path(p,S)$ is
{\sl OneOrMorePath($path(v,S)$)}.
\item If $p$ is a blank node that has exactly one value $v$ for {\tt sh:zeroOrOnePath} in
G and $v$ is not an ill-formed property path in $G$ then $path(p,S)$ is
{\sl ZeroOrOnePath($path(v,S)$)}.
\end{enumerate}

If an RDF term is not an ill-formed property path in an RDF graph $G$ then
it is an {\bf ASHACL property path} in $G$.

ASHACL property paths in same or different RDF graphs are {\bf path-equivalent} if
their path mappings in their graphs are the same.

\section{Validation Results and Validation Reports}

The results of ASHACL validation are RDF graphs that report on
the results the validation.

A {\bf results graph} is an RDF graph $G$ where every 
\href{https://www.w3.org/TR/rdf11-concepts/#section-rdf-graph}{\em node} in $G$ that has 
ASHACL type {\tt sh:ValidationResult} in $G$ meets the following conditions
\begin{itemize}
\item It has {\tt sh:ValidationResult} as a value for {\tt rdf:type} in $G$.
\item It has exactly one value for {\tt sh:focusNode} in $G$.
\item It has at most one value for {\tt sh:valueNode} in $G$.
\item It has at most one value for {\tt sh:resultPath}  in $G$ and this value
    if present is an ASHACL property path in $G$.
\item It has exactly one value for {\tt sh:sourceShape} in $G$.
\item It has exactly one value for {\tt sh:sourceConstraintComponent} in $G$.
\item It has exactly one value for {\tt sh:resultSeverity} in $G$.
\item Each of its values for {\tt sh:resultMessage} in $G$ is a
  \href{https://www.w3.org/TR/rdf11-concepts/#dfn-language-tagged-string}{\em language-tagged string} 
  and each has a different
  \href{https://www.w3.org/TR/rdf11-concepts/#dfn-language-tag}{\em language tag}. 
\item Each of its values for {\tt sh:detail} in $G$ has ASHACL type
    {\tt sh:ValidationResult} in $G$.
\item Each of its values for $p$ in $G$ where $p$ is a list-taking parameter is an
    ASHACL list in $G$.
\end{itemize}

\noindent
A {\bf validation result} from RDF term $f$, optional RDF term $v$, data graph
$D$, constraint $c$ that has kind $C$, and shape $s$ in shapes graph $S$
is a node in a results graph $G$ that meets the following conditions
\begin{itemize}
\item It has {\tt sh:ValidationResult} as a value for {\tt rdf:type} in $G$.
\item Its sole value for {\tt sh:focusNode} in $G$ is $f$.
\item Its sole value for {\tt sh:valueNode} in $G$ is $v$ if $v$ is present,
    otherwise it has no value for {\tt sh:valueNode} in $G$.
\item Its sole value for {\tt sh:resultPath} in $G$ is an ASHACL property path in $G$ that
    is path-equivalent to the value of $s$ for {\tt sh:path} in $S$ if there is one,
    otherwise it has no value for {\tt sh:resultPath} in $G$.
\item Its sole value for {\tt sh:sourceShape} in $G$ is $s$.
\item Its sole value for {\tt sh:sourceConstraintComponent} in $G$ is $C$.
\item Its sole value for {\tt sh:resultSeverity} in $G$ is the severity of $s$ in $S$.
\item Its values for {\tt sh:resultMessage} in $G$ include any values of $s$ for
    {\tt sh:message} in $S$.
\item Its values, if any, for $p$ in $G$ with $p$ a non-list-taking parameter of $C$
    and $X = \{ x | \left<p,x\right> \mbox{is a parameter value of $c$}\}$
    are the elements of $X$. 
\item Its values, if any, for $p$ in $G$ with $p$ a list-taking parameter of $C$
    and $X = \{ x | \left<p,x\right> \mbox{is a parameter value of $c$}\}$
    contain for each member $x$
    of $X$ an ASHACL list in $G$ whose elements in $G$ are the elements of
    $x$ in $S$ and no other values.
\end{itemize}

A {\bf results structure} is a results graph $R$ and a set of nodes from $R$ each
of which have ASHACL type {\tt sh:ValidationResult} in $R$.  The set of nodes is
called the {\bf top-level validation results} of the results structure.

A results structure {\bf contains} a top-level validation result from
$f$,$v$,$D$,$c$,$s$,$S$ (or $f$,$D$,$c$,$s$,$S$) if its graph contains a node
that is a validation result from $f$,$v$,$D$,$c$,$s$,$S$
($f$,$D$,$c$,$s$,$S$, respectively) and that node is an element of its top-level
validation results.

A {\bf validation report} $V$ for a results structure $R$ is any results graph
containing at least the RDF triples of $R$ plus a new blank node $n$ with
{\tt sh:ValidationReport} as a value for {\tt rdf:type} in $V$; {\tt "false"\string^\string^xsd:boolean} as
sole value for {\tt sh:conforms} in $V$ if $R$ has any top-level validation
results in $R$ and {\tt "true"\string^\string^xsd:boolean} otherwise; and whose set of values for
{\tt sh:result} in $V$ is the set of top-level validation results in $R$.

The {\bf combination} of a multiset of results structures is the union of the
graphs of the results structures and the union of their sets of top-level
validation results, provided that this is a results structure.

\noindent {\bf Note:} Combination does not use merging so that, for example,
validation results and ASHACL property paths do not have to be replicated.
Care does need to be taken when different results structures share blank
nodes so that invalid validation results or ASHACL property paths do not
result.

\noindent {\bf Note:} Validation reports are only required to contain
top-level validation results from results graphs, i.e., the results for
validating elements of the complete targets of shapes against the shape.
Implementations are encouraged to provide interfaces that also retain other
validation results and to use the property {\tt sh:detail} to link from a
validation result for a shape to validation results from subsidiary
conformance checking such as is required for constraints that have kind
{\tt sh:ShapeConstraintComponent}.

\section{Shapes}

\subsection{Shapes and Shapes Graphs}

The {\bf explicit shapes} of an RDF graph $G$, $ES(G)$, is the set of nodes in $G$ that
\begin{enumerate}
\item have {\tt sh:Shape} as an ASHACL type in $G$, or
\item are the subject of an RDF triple in $G$ with predicate one of
  the target predicates, {\tt sh:targetNode}, {\tt sh:targetClass},
  {\tt sh:targetSubjectsOf}, and {\tt sh:targetObjectsOf}.
\end{enumerate}
The {\bf shapes} of an RDF graph $G$, $S(G)$, is the smallest set of RDF terms
such that 
\begin{enumerate}
\item $ES(G) \subseteq S(G)$,
\item If $s \in S(G)$ and $\left< s, p, o \right>$ is a triple in $G$ wiith $p$
  a non-list-taking, shape-inducing parameter then $o \in S(G)$.
\item If $s \in S(G)$, $\left< s, p, l \right>$ is a triple in $G$ with $p$
  a list-taking, shape-inducing parameter and $l$ an ASHACL list in
  $G$, and $o$ is a mamber of $l$ in $G$ then  $o \in S(G)$.
\end{enumerate}

A {\bf shapes graph} is an RDF graph $G$ containing no ill-formed shapes in $G$.
{\bf Ill-formed shapes} are defined throughout the remainder of this document.

If $s$ is a shape in an RDF graph $G$ with more than one value for {\tt sh:severity}
in $G$ then $s$ is an {\bf ill-formed shape} in $G$.
The {\bf severity} of a shape in a shapes graph $S$ is its value for
{\tt sh:severity} in $S$, if any, otherwise {\tt sh:Violation}

If $s$ is a shape in an RDF graph $G$ with a value for {\tt sh:message} in $G$ that is
not either a language-tagged string or a
\href{https://www.w3.org/TR/rdf11-concepts/#dfn-literal}{\em literal} with
\href{https://www.w3.org/TR/rdf11-concepts/#dfn-datatype-iri}{\em datatype} {\tt xsd:string}
then $s$ is an {\bf ill-formed shape} in $G$.

If $s$ is a shape in an RDF graph $G$ with a value for {\tt sh:deactivated}
in $G$ that is not a literal with datatype {\tt xsd:boolean} then $s$ is an
{\bf ill-formed shape} in $G$.

\subsection{Constraint Components and Constraints}

A {\bf constraint component} is one of the IRIs so-described in the remainder of
this document.  Each constraint component has one or more {\bf mandatory
parameters}, each of which is a property.  Each constraint component has
zero or more {\bf optional parameters}, each of which is a property.  The
{\bf parameters} of a constraint component are its mandatory parameters plus
its optional parameters.  Some parameters of constraint components are
{\bf list-taking parameters}.  Parameters that are not list-taking
parameters are non-list-taking parameters.
Some parameters of constraint components are
{\bf shape-inducing parameters}. 

For a constraint component $C$ with mandatory parameters $p_1$, ..., $p_n$,
a shape s in a shapes graph $S$ has a {\bf constraint} that has kind $C$
with {\bf mandatory parameter values} $\left<p_1,v_1\right>$, \ldots,
$\left<p_n,v_n\right>$ in $S$ when $s$ has $v_i$ as a value for $p_i$ in
$S$.  If $s$ in $S$ has a constraint that has kind $C$ in $S$ then the {\bf
  optional parameter values} of the constraint in $S$ are all the
$\left<o_i,v_i\right>$ where $o_i$ is an optional parameter of $C$ and $s$
has $v_i$ as a value for $o_i$ in $S$.  The {\bf parameter values} of a
constraint are its mandatory parameter values plus its optional parameter
values.

\subsection{Recursive Shapes and Recursive Shapes Graphs}

A shape $s_1$ in an RDF graph $G$ {\bf refers} to shape $s_2$ in $G$ if it has $s_2$
as value for some non-list-taking, shape-inducing parameter of some
constraint component or $s_2$ as a member of the value for some list-taking,
shape-inducing parameter of some constraint component.

A shape in an RDF graph $G$ is {\bf recursive} in $G$ if it is related to itself
by the transitive closure of the refers relationship in $G$.  An RDF graph $G$
that contains a shape recursive in $G$ is {\bf recursive}.

\subsection{Targets}

If $s$ is a shape in an RDF graph $G$ with a value for {\tt sh:targetNode} in $G$ that
is not an IRI or literal then $s$ is an {\bf ill-formed shape} in $G$.
If $s$ is a shape in a shapes graph $S$ and $s$ has value t for {\tt sh:targetNode} in $S$
then $\{ t \}$ is a {\bf target} from any data graph for $s$ in $S$.

If $s$ is a shape in an RDF graph $G$ with a value for {\tt sh:targetClass} in $G$ that
is not an IRI then $s$ is an {\bf ill-formed shape} in $G$.
If $s$ is a shape in a shapes graph $S$ and $s$ has value $c$ for {\tt sh:targetClass} in
$S$ then the set of ASHACL instances of $c$ in a data graph $D$ is a {\bf target}
from $D$ for $s$ in $S$.

If $s$ is a shape in an RDF graph $G$ and $s$ is also an ASHACL instance of
{\tt rdfs:Class} in $G$ and $s$ is not an IRI then $s$ is an {\bf ill-formed shape} in $G$.
If $s$ is a shape in a shapes graph $S$ and $s$ is also an ASHACL instance of
{\tt rdfs:Class} in $S$ then the set of ASHACL instances of $s$ in a data graph $D$ is a
{\bf target} from $D$ for $s$ in $S$.

A shape in an RDF graph $G$ with a value for {\tt sh:targetSubjectsOf} in $G$ that is
not an IRI is an {\bf ill-formed} shape in $G$.
If $s$ is a shape in a shapes graph $S$ and $s$ has value $p$ for {\tt sh:targetSubjectsOf}
in $S$ then the set of nodes in a data graph $D$ that are subjects of RDF
triples in $D$ with predicate $p$ is a {\bf target} from $D$ for $s$ in $S$.

A shape in an RDF graph $G$ with a value for {\tt sh:targetObjectsOf} in $G$ that is
not an IRI is an {\bf ill-formed shape} in $G$.
If $s$ is a shape in a shapes graph $S$ and $s$ has value $p$ for {\tt sh:targetObjectsOf}
in $S$ then the set of nodes in a data graph $D$ that are objects of RDF triples
in $D$ with predicate $p$ is a {\bf target} from $D$ for $s$ in $S$.

The {\bf complete targets} from a data graph $D$ for a shape $s$ in a shapes graph
$S$ is the union of the targets from $D$ for $s$ in $S$ as defined above.

\subsection{Value Nodes}

A shape in an RDF graph $G$ with a value for {\tt sh:path} in $G$ that is an
ill-formed property path in $G$ is an {\bf ill-formed shape} in $G$.  A shape in
an RDF graph $G$ with more than one value for {\tt sh:path} in $G$ is an {\bf ill-formed
shape} in $G$.

Given $f$ an RDF term, $D$ a data graph, and $s$ a shape in $S$ a shapes graph
the {\bf value nodes} of $f$ with $D$ for $s$ in $S$ is the set containing
the values of $f$ for $path(p,S)$ in $D$ if $s$ has $p$ as value for {\tt sh:path} in $S$,
or just $f$ if $s$ has no value for {\tt sh:path} in $S$.

\section{Validation and Conformance Checking}

Validation is the main relationship defined in ASHACL, usually producing a
validation report containing the results of the validation.  There are
generally many different possible validation reports for a particular
validation.  Conformance checking is a simplified version of validation,
usually producing a boolean result.

Validation and conformance checking can result in a failure.  For example, a
particular ASHACL processor might allow recursive shapes but result in a failure
if it detects a loop.  Failure can also result from resource
exhaustion.  Failures are reported through implementation-specific
channels.

\paragraph*{\bf Validating a data graph against a shapes graph:}
Given $G$ a data graph and $S$ a shapes graph, a {\bf results structure} for the
validation of $G$ against $S$ is a combination of some multiset
$\{\!\!\{ R_1, \ldots, R_n\}\!\!\}$
where
$\{ s_1, \ldots, s_n \}$ is the set of shapes in $S$ and $R_i$ is a results
structure for the validation of $G$ against $s_i$ in $S$.  A {\bf validation report}
for the validation of $G$ against $S$ is a validation report for some results
structure for the validation of $G$ against $S$.  A data graph $D$ {\bf conforms} to
a shapes graph $S$ if and only if there is a results structure for the validation
of $D$ against $S$ that contains no top-level validation results.

\paragraph*{\bf Validating a data graph against a shape in a shapes graph:}
Given $G$ a data graph and $s$ a shape in $S$ a shapes graph, a {\bf results
structure} for the validation of $G$ against $s$ in $S$ is a combination of some
multiset
$\{\!\!\{ R_1, \ldots, R_n \}\!\!\}$ where $\{ t_1, \ldots, t_n \}$ is the complete targets
from $G$ for $s$ in $S$ and $R_i$ is a results structure for the validation of ti
using $G$ against $s$ in $S$.  A data graph $D$ {\bf conforms} to a shape $s$ in a
shapes graph $S$ if and only if there is a results structure for the validation
of $D$ against $s$ in $S$ that contains no top-level validation results.

\paragraph*{\bf Validating an RDF term using a data graph against a shape in a shapes graph:}
Given $f$ an RDF term, $D$ a data graph, and $s$ a shape in a shapes graph $S$ a
{\bf results structure} for the validation of $f$ using $G$ against $s$ in $S$ is a
combination of some multiset
$\{\!\!\{ R_1, \ldots, R_n \}\!\!\}$ where $\{ c_1, \ldots, c_n \}$ is the
constraints of $s$ in $S$, $R_i$ is an element of $results(f,V,D,ci,s,S)$, and $v$ is
the value nodes of $f$ with $D$ for $s$ in $S$, provided that $s$ does not have a
value for {\tt sh:deactivated} in $S$ that is a 
literal
with datatype {\tt xsd:boolean} and whose 
\href{https://www.w3.org/TR/rdf11-concepts/#dfn-literal-value}{\em literal value} 
is {\tt true}.  If $s$ does have such a value for
{\tt sh:deactivated} in $S$ then any {\bf results structure} for the validation of f
using $G$ against $s$ in $S$ has no top-level validation results.  An RDF term f
and a data graph $D$ {\bf conform} to a shape $s$ in a shapes graph $S$ if and only
if there is a results structure for the validation of $f$ using $D$ against $s$ in
S that contains no top-level validation results.

\medskip

\noindent {\bf Note:} Although there can be multiple possible results
structures for a particular validation, if any results structure for the
validation has no top-level validation results they will all have no
top-level validation results.

\section{Constraint1 Components and Validation}

This section defines the ASHACL constraint components along with their
parameters and how they participate in validation.

\medskip

Given an RDF term $f$, a set of RDF terms $v$, a data graph $D$, a constraint $c$,
and a shape $s$ in a shapes graph $S$ the possible {\bf results of validating} f
and $v$ with $D$ against $c$ in $S$, $results(f,V,D,c,s,S)$, is a set of results
structures as defined in the rest of this section.

\paragraph*{{\tt sh:ClassConstraintComponent}} is a {\bf constraint component} with mandatory
parameter {\tt sh:class}.
A shape in an RDF graph $G$ with a value for {\tt sh:class} in $G$ that is not
an IRI is an {\bf ill-formed shape} in $G$.

Each results structure in $results(f,V,D,c,s,S)$ where $c$ is a constraint
that has kind
{\tt sh:ClassConstraintComponent} and that has mandatory parameter values $\left<\mbox{\tt sh:class},x\right>$
contains a different top-level validation result from $f$,$v$,$D$,\discretionary{}{}{}$c$,$s$,$S$
for each $v$ in $V$ that is not an ASHACL instance of $x$ in $D$,
and no other top-level validation results.

\paragraph*{{\tt sh:DatatypeConstraintComponent}} is a {\bf constraint component} with mandatory
parameter {\tt sh:datatype}.
A shape in an RDF graph $G$ with a value for {\tt sh:datatype} in $G$ that is not
an IRI is an {\bf ill-formed shape} in $G$.

Each results structure in $results(f,V,D,c,s,S)$ where $c$ is a constraint
that has kind
{\tt sh:DatatypeConstraintComponent} and that has mandatory parameter values $\left<\mbox{\tt sh:datatype},x\right>$
contains a different top-level validation result from $f$,$v$,$D$,$c$,$s$,$S$
for each $v$ in $V$ that is not a literal with datatype $x$,
and no other top-level validation results.

\paragraph*{{\tt sh:NodeKindConstraintComponent}} is a {\bf constraint component} with mandatory
parameter {\tt sh:nodeKind}.
A shape in an RDF graph $G$ with a value for {\tt sh:nodeKind} in $G$ that is not
one of {\tt sh:BlankNode}, {\tt sh:IRI}, or {\tt sh:Literal} is an {\bf ill-formed shape} in $G$.

Each results structure in $results(f,V,D,c,s,S)$ where $c$ is a constraint
that has kind
{\tt sh:NodeKindConstraintComponent} and that has mandatory parameter values $\left<\mbox{\tt sh:nodeKind},x\right>$
contains a different top-level validation result from $f$,$v$,$D$,$c$,$s$,$S$
for each $v$ in $V$ that is not a blank node if $x$ is {\tt sh:BlankNode}, 
not an IRI if $x$ is {\tt sh:IRI}, or
not a literal if $x$ is {\tt sh:Literal},
and no other top-level validation results.

\paragraph*{{\tt sh:MinCountConstraintComponent}} is a {\bf constraint component} with mandatory
parameter {\tt sh:minCount}.
A shape in an RDF graph $G$ with a value for {\tt sh:minCount} in $G$ that is not a
literal
with datatype {\tt xsd:integer} is an {\bf ill-formed shape} in $G$.

Each results structure in $results(f,V,D,c,s,S)$ where $c$ is a constraint
that has kind
{\tt sh:MinCountConstraintComponent} and that has mandatory
parameter values $\left<\mbox{\tt sh:minCount},i\right>$ contains
a single top-level validation result from $f$,$D$,$c$,$s$,$S$ and no other top-level
validation results if the cardinality of $v$ is less than the 
literal value of $i$,
and no top-level validation results otherwise.

\paragraph*{{\tt sh:MaxCountConstraintComponent}} is a {\bf constraint component} with mandatory
parameter {\tt sh:maxCount}.
A shape in an RDF graph $G$ with a value for {\tt sh:maxCount} in $G$ that is not a
literal
with datatype {\tt xsd:integer} is an {\bf ill-formed shape} in $G$.

Each results structure in $results(f,V,D,c,s,S)$ where $c$ is a constraint
that has kind
{\tt sh:MaxCountConstraintComponent} and that has mandatory
parameter values $\left<\mbox{\tt sh:maxCount},i\right>$ contains
a single top-level validation result from  $f$,$D$,$c$,$s$,$S$ and no other top-level
validation results if the cardinality of $v$ is greater than the 
literal value of $i$,
and no top-level validation results otherwise.

\paragraph*{{\tt sh:MinExclusiveConstraintComponent}} is a {\bf constraint component} with
mandatory parameter {\tt sh:minExclusive}.
A shape in an RDF graph $G$ with a value for {\tt sh:minExclusive} in $G$ that is not
suitable for use in the 
\href{https://www.w3.org/TR/sparql11-query/#OperatorMapping}{\em SPARQL operator {\tt <}} 
is an {\bf ill-formed shape} in $G$.

Each results structure in $results(f,V,D,c,s,S)$ where $c$ is a constraint
that has kind
{\tt sh:MinExclusiveConstraintComponent} and that has mandatory parameter values $\left<\mbox{\tt sh:minExclusive},i\right>$
contains a different top-level validation result from $f$,$v$,$D$,$c$,$s$,$S$
for each $v$ in $V$ for which $i {\tt <} v$ returns false or produces an error in SPARQL,
and no other top-level validation results.

\paragraph*{{\tt sh:MinInclusiveConstraintComponent}} is a {\bf constraint component} with
mandatory parameter {\tt sh:minInclusive}.
A shape in an RDF graph $G$ with a value for {\tt sh:minInclusive} in $G$ that is not
suitable for use in the 
\href{https://www.w3.org/TR/sparql11-query/#OperatorMapping}{\em SPARQL operator {\tt <=}} 
is an {\bf ill-formed shape} in $G$.

Each results structure in $results(f,V,D,c,s,S)$ where $c$ is a constraint
that has kind
{\tt sh:MinInclusiveConstraintComponent} and that has mandatory parameter values $\left<\mbox{\tt sh:minInclusive},i\right>$
contains different top-level validation result from $f$,$v$,$D$,$c$,$s$,$S$
for each $v$ in $V$ for which $i {\tt <=} v$ returns false or produces an error in SPARQL,
and no other top-level validation results.

\paragraph*{{\tt sh:MaxExclusiveConstraintComponent}} is a {\bf constraint component} with
mandatory parameter {\tt sh:maxExclusive}.
A shape in an RDF graph $G$ with a value for {\tt sh:maxExclusive} in $G$ that is not
suitable for use in the 
\href{https://www.w3.org/TR/sparql11-query/#OperatorMapping}{\em SPARQL operator {\tt >}} 
is an {\bf ill-formed shape} in $G$.

Each results structure in $results(f,V,D,c,s,S)$ where $c$ is a constraint
that has kind
{\tt sh:MaxExclusiveConstraintComponent} and that has mandatory parameter values $\left<\mbox{\tt sh:maxExclusive},i\right>$
contains a different top-level validation result from $f$,$v$,$D$,$c$,$s$,$S$
for each $v$ in $V$ for which $i {\tt >} v$ returns false or produces an error in SPARQL,
and no other top-level validation results.

\paragraph*{{\tt sh:MaxInclusiveConstraintComponent}} is a {\bf constraint component} with
mandatory parameter {\tt sh:maxInclusive}.
A shape in an RDF graph $G$ with a value for {\tt sh:maxInclusive} in $G$ that is not
suitable for use in the 
\href{https://www.w3.org/TR/sparql11-query/#OperatorMapping}{\em SPARQL operator {\tt >=}} 
is an {\bf ill-formed shape} in $G$.

Each results structure in $results(f,V,D,c,s,S)$ where $c$ is a constraint
that has kind
{\tt sh:MaxInclusiveConstraintComponent} and that has mandatory parameter values $\left<\mbox{\tt sh:maxInclusive},i\right>$
contains a different top-level validation result from $f$,$v$,$D$,$c$,$s$,$S$
for each $v$ in $V$ for which $i {\tt >=} v$ returns false or produces an error in SPARQL,
and no other top-level validation results.

\paragraph*{{\tt sh:MinLengthConstraintComponent}} is a {\bf constraint component} with
mandatory parameter {\tt sh:minLength}.
A shape in an RDF graph $G$ with a value for {\tt sh:minLength} in $G$ that is not
a literal with datatype {\tt xsd:integer}
is an {\bf ill-formed shape} in $G$.

Each results structure in $results(f,V,D,c,s,S)$ where $c$ is a constraint
that has kind
{\tt sh:MinLengthConstraintComponent} and that has mandatory parameter values $\left<\mbox{\tt sh:minLength},i\right>$
contains a different top-level validation result from $f$,$v$,$D$,$c$,$s$,$S$
for each $v$ in $V$ that is a blank node or where the length of its string
representation (as defined by the  
\href{https://www.w3.org/TR/sparql11-query/#func-str}{\em SPARQL {\tt str} function}) 
is less than the literal value of $i$,
and no other top-level validation results.

\paragraph*{{\tt sh:MaxLengthConstraintComponent}} is a {\bf constraint component} with
mandatory parameter {\tt sh:maxLength}.
A shape in an RDF graph $G$ with a value for {\tt sh:maxLength} in $G$ that is not
a literal with datatype {\tt xsd:integer}
is an {\bf ill-formed shape} in $G$.

Each results structure in $results(f,V,D,c,s,S)$ where $c$ is a constraint
that has kind
{\tt sh:MaxLengthConstraintComponent} and that has mandatory parameter values $\left<\mbox{\tt sh:maxLength},i\right>$
contains a different top-level validation result from $f$,$v$,$D$,$c$,$s$,$S$
for each $v$ in $V$
that is a blank node or where the length of its string representation (as defined by the  
\href{https://www.w3.org/TR/sparql11-query/#func-str}{\em SPARQL {\tt str} function}) 
is greater than the literal value of $i$,
and no other top-level validation results.

\paragraph*{{\tt sh:PatternConstraintComponent}} is a {\bf constraint component} with mandatory
parameter {\tt sh:pattern} and optional parameter {\tt sh:flags}.
A shape in an RDF graph $G$ with a value for {\tt sh:pattern} or {\tt sh:flags} in $G$ that
is not a literal with datatype {\tt xsd:string} is an {\bf ill-formed shape} in $G$.
A shape in an RDF graph $G$ with more than one value for {\tt sh:pattern} or
{\tt sh:flags} in $G$ is an {\bf ill-formed shape} in $G$.

Each results structure in $results(f,V,D,c,s,S)$ where $c$ is a constraint
that has kind
{\tt sh:PatternConstraintComponent} and that has mandatory parameter values $\left<\mbox{\tt sh:pattern},r\right>$
contains a different top-level validation result from $f$,$v$,$D$,$c$,$s$,$S$
for each $v$ in $V$
that is either a blank node or where the string representation (as
defined by the 
\href{https://www.w3.org/TR/sparql11-query/#func-str}{\em SPARQL {\tt str} function}) 
does not match the given regular
expression (as defined by the
\href{https://www.w3.org/TR/sparql11-query/#func-regex}{\em SPARQL {\tt REGEX} function}),
and no other top-level validation results.
If the constraint has optional parameter values $\left<\mbox{\tt sh:flags},flags\right>$
then flags is used as the third argument of the 
\href{https://www.w3.org/TR/sparql11-query/#func-regex}{\em SPARQL {\tt REGEX} function},
otherwise the third argument is the empty string.

\paragraph*{{\tt sh:StemConstraintComponent}} is a {\bf constraint component} with mandatory
parameter {\tt sh:stem}
A shape in an RDF graph $G$ with a value for {\tt sh:stem} that
is not a literal with datatype {\tt xsd:string} is an {\bf ill-formed shape} in $G$.

Each results structure in $results(f,V,D,c,s,S)$ where $c$ is a constraint
that has kind
{\tt sh:StemConstraintComponent} and that has mandatory parameter values $\left<\mbox{\tt sh:stem},i\right>$
contains a different top-level validation result from $f$,$v$,$D$,$c$,$s$,$S$
for each $v$ in $V$
that is not an IRI or where its string representation (as defined by the  
\href{https://www.w3.org/TR/sparql11-query/#func-str}{\em SPARQL {\tt str} function}) 
does not
start with (as defined by the 
\href{https://www.w3.org/TR/2013/REC-sparql11-query-20130321/#func-strstarts}{\em SPARQL {\tt STRSTARTS} function})
the literal value of $i$,
and no other top-level validation results.

\paragraph*{{\tt sh:LanguageInConstraintComponent}} is a {\bf constraint component} with mandatory
list-taking parameter {\tt sh:languageIn}.
A shape in an RDF graph $G$ with a value for {\tt sh:languageIn} in $G$ that
is not an ASHACL list in $G$ whose members in $G$ are all literals with
datatype {\tt xsd:string} is an {\bf ill-formed shape} in $G$.

Each results structure in $results(f,V,D,c,s,S)$ where $c$ is a constraint
that has kind
{\tt sh:LanguageInConstraintComponent} and that has mandatory parameter values $\left<\mbox{\tt sh:languageIn},l\right>$
contains a different top-level validation result from $f$,$v$,$D$,$c$,$s$,$S$
for each $v$ in $V$
that is not a language-tagged string whose language tag matches the literal
value of one of elements of $l$
as defined by the 
\href{https://www.w3.org/TR/sparql11-query/#func-langMatches}{SPARQL {\tt langMatches} function},
and no other top-level validation results.

\paragraph*{{\tt sh:UniqueLangConstraintComponent}} is a {\bf constraint component} with
mandatory parameter {\tt sh:uniqueLang}.
A shape in an RDF graph $G$ with a value for {\tt sh:uniqueLang} in $G$ that is not a
literal with datatype {\tt xsd:boolean} whose
literal value
is {\tt true} is an {\bf ill-formed shape} in $G$.

Each results structure in $results(f,V,D,c,s,S)$ where $c$ is a constraint
that has kind
{\tt sh:UniqueLangConstraintComponent} and that has mandatory parameter values $\left<\mbox{\tt sh:uniqueLang},t\right>$
for $t$ a literal with literal value {\tt true}
contains a different top-level validation result from $f$,$v$,$D$,$c$,$s$,$S$
for each $v$ in $V$ that is is a
language-tagged string
with the same
language tag
as some other element of $V$,
and no other top-level validation results.

\paragraph*{{\tt sh:EqualsConstraintComponent}} is a {\bf constraint component} with mandatory
parameter {\tt sh:equals}.
A shape in an RDF graph $G$ with a value for {\tt sh:equals} in $G$ that is not an IRI
is an {\bf ill-formed shape} in $G$.

Each results structure in $results(f,V,D,c,s,S)$ where $c$ is a constraint
that has kind
{\tt sh:EqualsConstraintComponent} and that has mandatory parameter values $\left<\mbox{\tt sh:equals},p\right>$
contains a different top-level validation result from $f$,$v$,$D$,$c$,$s$,$S$
for each $v$ in $V$ that is not a value of $f$ for $p$ in $D$
and for each $v$ a value of $f$ for $p$ in $D$ that is not in $V$,
and no other top-level validation results.

\paragraph*{{\tt sh:DisjointConstraintComponent}} is a {\bf constraint component} with mandatory
parameter {\tt sh:disjoint}.
A shape in an RDF graph $G$ with a value for {\tt sh:disjoint} in $G$ that is not an
IRI is an {\bf ill-formed shape} in $G$.

Each results structure in $results(f,V,D,c,s,S)$ where $c$ is a constraint
that has kind
{\tt sh:DisjointConstraintComponent} and that has mandatory parameter values $\left<\mbox{\tt sh:disjoint},p\right>$
contains a different top-level validation result from $f$,$v$,$D$,$c$,$s$,$S$
for each $v$ in $V$ that is also a value of $f$ for $p$ in $D$,
and no other top-level validation results.

\paragraph*{{\tt sh:LessThanConstraintComponent}} is a {\bf constraint component} with mandatory
parameter {\tt sh:lessThan}.
A shape in an RDF graph $G$ with a value for {\tt sh:lessThan} in $G$ that is not 
an IRI
is an {\bf ill-formed shape} in $G$.

Each results structure in $results(f,V,D,c,s,S)$ where $c$ is a constraint
that has kind
{\tt sh:LessThanConstraintComponent} and that has mandatory parameter values $\left<\mbox{\tt sh:lessThan},p\right>$
contains a different top-level validation result from $f$,$v$,$D$,$c$,$s$,$S$
for each $v$ in $V$ that does not compare as less than
each value of $f$ for $p$ in $G$ using 
\href{https://www.w3.org/TR/sparql11-query/#OperatorMapping}{\em SPARQL operator {\tt <}},
and no other top-level validation results.

\noindent {\bf Note:}  A comparison that produces an error counts as not comparing as less than.

\paragraph*{{\tt sh:LessThanOrEqualsConstraintComponent}} is a {\bf constraint component} with mandatory
parameter {\tt sh:lessThanOrEquals}.
A shape in an RDF graph $G$ with a value for {\tt sh:lessThanOrEquals} in $G$ that is not 
an IRI
is an {\bf ill-formed shape} in $G$.

Each results structure in $results(f,V,D,c,s,S)$ where $c$ is a constraint
that has kind
{\tt sh:LessThanOrEqualsConstraintComponent} and that has mandatory parameter values $\left<\mbox{\tt sh:lessThanOrEquals},p\right>$
contains a different top-level validation result from $f$,$v$,$D$,$c$,$s$,$S$
for each $v$ in $V$ that does not compare as less than or equal
each value of $f$ for $p$ in $G$ using 
\href{https://www.w3.org/TR/sparql11-query/#OperatorMapping}{\em SPARQL operator {\tt <=}},
and no other top-level validation results.

\noindent {\bf Note:}  A comparison that produces an error counts as not comparing as
less than or equals.

\paragraph*{{\tt sh:ShapeConstraintComponent}} is a {\bf constraint component} with mandatory
shape-inducing parameter {\tt sh:shape}.
A shape in an RDF graph $G$ with a value for {\tt sh:shape} in $G$ that is not an IRI
or blank node is an {\bf ill-formed shape} in $G$.

Each results structure in $results(f,V,D,c,s,S)$ where $s$ is not recursive and $c$
is a constraint that has kind {\tt sh:ShapeConstraintComponent} and that has
mandatory parameter values $\left<\mbox{\tt sh:shape},x\right>$
contains a different top-level validation result from $f$,$v$,$D$,$c$,$s$,$S$ for each $v$
in $v$ where $v$ and $D$ do not conform to $x$ in $S$,
and no other top-level validation results.

\paragraph*{{\tt sh:NotConstraintComponent}} is a {\bf constraint component} with mandatory
shape-inducing parameter {\tt sh:not}.
A shape in an RDF graph $G$ with a value for {\tt sh:not} in $G$ that is not an IRI
or blank node is an {\bf ill-formed shape} in $G$.

Each results structure in $results(f,V,D,c,s,S)$ where $s$ is not recursive and $c$
is a constraint that has kind {\tt sh:NotConstraintComponent} and that has
mandatory parameter values $\left<\mbox{\tt sh:not},x\right>$
contains a different top-level validation result from $f$,$v$,$D$,$c$,$s$,$S$ for each $v$
in $v$ where $v$ and $D$ do conform to $x$ in $S$,
and no other top-level validation results.

\paragraph*{{\tt sh:AndConstraintComponent}} is a {\bf constraint component} with mandatory
list-taking, shape-inducing parameter {\tt sh:and}.
A shape in an RDF graph $G$ with a value for {\tt sh:and} in $G$ that is not an ASHACL
list in $G$ whose members in $G$ are all IRIs or blank nodes in $G$ is an
{\bf ill-formed shape} in $G$.

Each results structure in $results(f,V,D,c,s,S)$ where $s$ is not recursive
and $c$ is a constraint that has kind {\tt sh:AndConstraintComponent} and that has mandatory
parameter values $\left<\mbox{\tt sh:and},l\right>$ contains a different top-level
validation result from $f$,$v$,$D$,$c$,$s$,$S$ for each $v$ in $V$ where $v$
and $D$ do not conform to each element of $l$ in $S$,
and no other top-level validation results.

\paragraph*{{\tt sh:OrConstraintComponent}} is a {\bf constraint component} with mandatory
list-taking, shape-inducing parameter {\tt sh:or}.
A shape in an RDF graph $G$ with a value for {\tt sh:or} in $G$ that is not an ASHACL
list in $G$ whose members in $G$ are all IRIs or blank nodes in $G$ is an
{\bf ill-formed shape} in $G$.

Each results structure in $results(f,V,D,c,s,S)$ where $s$ is not recursive
and $c$ is a constraint that has kind {\tt sh:OrConstraintComponent} and that has mandatory
parameter values $\left<\mbox{\tt sh:or},l\right>$ contains a different top-level
validation result from $f$,$v$,$D$,$c$,$s$,$S$ for each $v$ in $V$ where $v$
and $D$ do not conform to any element of $l$ in $S$,
and no other top-level validation results.

\paragraph*{{\tt sh:QualifiedValueShapeConstraintComponent}} is a {\bf constraint component}
with mandatory shape-inducing parameter {\tt sh:qualifiedValueShape} and optional parameters
{\tt sh:qualifiedMinCount} and {\tt sh:qualifiedMaxCount}
A shape in an RDF graph $G$ with a value for {\tt sh:qualifiedValueShape} in $G$ that
is not an IRI or blank node is an {\bf ill-formed shape} in $G$.
A shape in an RDF graph $G$ with a value for {\tt sh:qualifiedMinCount} or
{\tt sh:qualifiedMaxCount} in $G$ that is not a literal with datatype {\tt xsd:integer} is
an {\bf ill-formed shape} in $G$.
A shape in an RDF graph $G$ with more than one value for
{\tt sh:qualifiedValueShape}, {\tt sh:qualifiedMinCount}, or {\tt sh:qualifiedMaxCount} in $G$
is an {\bf ill-formed shape} in $G$.
A shape in an RDF graph $G$ that has a value for {\tt sh:qualifiedValueShape} in $G$
but no value for either {\tt sh:qualifiedMinCount} or
{\tt sh:qualifiedMaxCount} in $G$ is an {\bf ill-formed shape} in $G$.

Each results structure in $results(f,V,D,c,s,S)$ where $s$ is not recursive and $c$
is a constraint that has kind {\tt
  sh:QualifiedValueShapeConstraintComponent} and that has mandatory parameter
values $\left<\mbox{\tt sh:qualifiedValueShape},x\right>$
contains a single top-level validation result from $f$,$D$,$c$,$s$,$S$ 
and no other top-level validation results
if the number of elements $v$ of $v$ where where $v$ and $D$ conform to
$x$ in $S$ is less than the
literal value
of $m$ if $\left<\mbox{\tt sh:qualifiedMinCount},m\right>$ is in the
optional parameters of the constraint or is greater than the 
literal value
$\left<\mbox{\tt sh:qualifiedMaxCount},n\right>$ is in the optional parameters of the constraint,
and no top-level validation results otherwise.

\paragraph*{{\tt sh:ClosedConstraintComponent}} is a {\bf constraint component} with mandatory
parameter {\tt sh:closed} and optional list-taking parameter {\tt sh:ignoredProperties}
A shape in an RDF graph $G$ with a value for {\tt sh:closed} in $G$ that is not a
literal with datatype {\tt xsd:boolean} whose value is {\tt true} is an {\bf ill-formed
shape} in $G$.
A shape in an RDF graph $G$ with a value for {\tt sh:ignoredProperties} in $G$ that is
not an ASHACL list in $G$ whose members in $G$ are all IRIs is an {\bf ill-formed
shape} in $G$.
A shape in an RDF graph $G$ with more than one value for {\tt sh:closed} or
{\tt sh:ignoredProperties} in $G$ is an {\bf ill-formed shape} in $G$.

Each results structure in $results(f,V,D,c,s,S)$ where $c$ is a constraint
that has kind
{\tt sh:ClosedConstraintComponent} and that has mandatory parameter values $\left<\mbox{\tt sh:closed},t\right>$
contains a different top-level validation result from $f$,$v$,$D$,$c$,$s$,$S$
for each RDF triple in $D$ with subject $f$
whose predicate is
not a value of some value of $s$ for {\tt sh:shape} in $D$ for {\tt sh:path} in $D$
and not a member of $i$ in $D$ if the constraint has
optional parameter values $\left<\mbox{\tt sh:ignoredProperties},i\right>$,
and no other top-level validation results.

\paragraph*{{\tt sh:HasValueConstraintComponent}} is a {\bf constraint component} with mandatory
parameter {\tt sh:hasValue}.

Each results structure in $results(f,V,D,c,s,S)$ where $c$ is a constraint
that has kind
{\tt sh:HasValueConstraintComponent} and that has mandatory parameter values
$\left<\mbox{\tt sh:hasValue},x\right>$ contains
a single top-level validation result from  $f$,$D$,$c$,$s$,$S$ 
and no other top-level validation results if $x$ is not in $V$,
and no top-level validation results otherwise.

\paragraph*{{\tt sh:InConstraintComponent}} is a {\bf constraint component} with mandatory
list-taking parameter {\tt sh:in}
A shape in an RDF graph $G$ with a value for {\tt sh:in} in $G$ that is not an ASHACL
list in $G$ whose members in $G$ are all IRIs or blank nodes in $G$ is an
{\bf ill-formed shape} in $G$.

Each results structure in $results(f,V,D,c,s,S)$ where $c$ is a constraint
that has kind
{\tt sh:InConstraintComponent} and that has mandatory parameter values $\left<\mbox{\tt sh:in},x\right>$ contains
a different top-level validation result from $f$,$v$,$D$,$c$,$s$,$S$ for each $v$ in $V$
where $v$ is not a member of $x$ in $S$,
and no other top-level validation results.

\section{ASHACL Processors and Conformance}

The keywords MAY, MUST, MUST NOT, and SHOULD are to be interpreted in this
section as described in \href{https://tools.ietf.org/html/rfc2119}{RFC2119} \cite{rfc2119}.

An ASHACL processor MUST provide an interface that takes two RDF graphs---a
data graph and a potential shapes graph.

If the potential shapes graph is not a shapes graph the processor MUST
report a failure.  If the potential shapes graph is recursive the behaviour
of the processor is not fully defined here; the processor MAY report a
failure if the potential shapes graph is recursive or MAY extend this
specification.  The behavior of an ASHACL processor MUST satisfy all the
requirements in this document regardless of whether the the potential
shapes graph is recursive or not.

Unless the ASHACL processor reports a failure, it MUST either return a
representation, or a location from which a representation can be retrieved,
for an RDF graph that is a validation report for the validation of the data
graph against the shapes graph.  The representation can be a document in a
syntax for RDF graphs, such as Turtle, or an RDF graph stored in a graph
repository.

An ASHACL processor SHOULD provide a version of the above interface where
the representation of the validation report is a file containing a Turtle
document.

An ASHACL processor SHOULD provide an interface similar to the first
interface above
except that the return value of the interface 
is a boolean indicating whether the data graph conforms to the 
shapes graph.

An ASHACL processor SHOULD provide versions of the above interfaces where the
potential shapes graph is formed by starting with an initial graph and
merging in other graphs accessible at locations that are values of RDF
triples with predicate {\tt owl:imports} in the potential shapes graph being
formed.

An ASHACL processor SHOULD provide versions of the above
interfaces where the potential shapes graph is formed by accessing
locations that are objects of RDF triples with predicate {\tt
  sh:shapesGraph} in the data graph and merging the RDF graphs available at
these locations into an initial graph or potential shapes graph.

An ASHACL processor SHOULD provide versions of the above interfaces
where the data graph is the result of using {\tt owl:imports}
as above starting with some initial data graph 
or 
where the data graph is the result of performing inference according to the
inferences required in one or more SPARQL entailment regimes
\cite{sparql-entailment} specified as the objects of RDF triples in the shapes
graph with predicate {\tt sh:entailment}
starting with an initial data graph.

An ASHACL processor SHOULD provide two extra versions of each interface
above that it provides.
The first of these takes in addition an IRI and validates or checks
conformance with respect to the data graph and the IRI as a shape in the
shapes graph.
The second of these takes in addition also an IRI or RDF literal and
validates or checks conformance with respect to the IRI or RDF literal using
the data graph and the IRI as a shape in the shapes graph.

If an ASHACL processor accesses any persistent representation of any graph
that contributes to the data graph or the potential shapes graph the ASHACL
processor MUST NOT change these representations.  An ASHACL processor MAY
create a persistent representation of a validation report that it constructs
so long as the ASHACL processor does not not violate the previous
requirement.

\section{Conclusion}

It is easier to fix SHACL Core, which has been done here, than it is to fix
SHACL-SPARQL, which requires at least a suitable definition of pre-binding.
A workable specification of SHACL-SPARQL is thus deferred to future work.

\section{Changes from previous version}

This version of the document extends the conditions for ill-formed property
paths include blank nodes that have a value for {\tt rdf:rest} and a value
for one of the ASHACL properties that are used to encode SPARQL property
path expressions.  This version of the document limits the shapes in a
shapes graph to those nodes in the graph that are SHACL instances of {\tt
  sh:Shape}, have a value for any of the target predicates, or are used as
shapes in these nodes.

\bibliography{sparql}

\begin{thebibliography}{1}

\bibitem{rfc2119}
S.~Brander.
\newblock Key words for use in {RFC}s to indicate requirement levels.
\newblock IETF Request for Comments,
  \url{https://www.ietf.org/rfc/rfc2119.txt}, March 1997.

\bibitem{rdf11-concepts}
Richard Cyganiak and David Wood.
\newblock {RDF} 1.1 concepts and abstract syntax.
\newblock {W3C} Last Call Working Draft,
  \url{http://www.w3.org/TR/rdf11-concepts}, 23 July 2013.

\bibitem{sparql-entailment}
Birte Glimm and Chimezie Ogbuji.
\newblock {SPARQL} 1.1 entailment regimes.
\newblock W3C Rec.,
  \url{https://www.w3.org/TR/2013/REC-sparql11-entailment-20130321/}, 21 March
  2013.

\bibitem{sparql}
Steve Harris and Andy Seaborne.
\newblock {SPARQL} 1.1 query language.
\newblock W3C Rec.,
  \url{https://www.w3.org/TR/2013/REC-sparql11-query-20130321/}, 21 March 2013.

\bibitem{shacl}
Holger Knublauch and Dimitris Kontokostas.
\newblock Shapes constraint language ({SHACL}).
\newblock W3C Working Draft,
  \url{https://www.w3.org/TR/2017/WD-shacl-20170202/}, 02 February 2017.

\bibitem{turtle-1.1}
Eric Prud'hommeaux and Gavin Carothers.
\newblock {RDF} 1.1 turtle: Terse {RDF} triple language.
\newblock {W3C} Recommendation, \url{http://www.w3.org/TR/turtle/}, February
  2014.

\end{thebibliography}
\bibliographystyle{plain}

\end{document}